\documentclass[10pt,twocolumn,letterpaper]{article}

\usepackage{iccv}
\usepackage{times}
\usepackage{epsfig}
\usepackage{graphicx}
\usepackage{amsmath}
\usepackage{amssymb}
\usepackage{subfigure}
\usepackage{tabularx}
\usepackage{arydshln}
\usepackage{array}
\newcolumntype{C}[1]{>{\centering\let\newline\\\arraybackslash\hspace{0pt}}p{#1}}

\usepackage[pagebackref=true,breaklinks=true,letterpaper=true,colorlinks,bookmarks=false]{hyperref}

\iccvfinalcopy 

\def\iccvPaperID{325} 

\ificcvfinal\fi

\begin{document}

\title{Learning Deconvolution Network for Semantic Segmentation}

\author{Hyeonwoo Noh\hspace{1.7cm}Seunghoon Hong\hspace{1.7cm}Bohyung Han\\
Department of Computer Science and Engineering, POSTECH, Korea\\
{\tt\small \{hyeonwoonoh\_,maga33,bhhan\}@postech.ac.kr}}

\maketitle

\begin{abstract}
We propose a novel semantic segmentation algorithm by learning a deconvolution network.
We learn the network on top of the convolutional layers adopted from VGG 16-layer net. 
The deconvolution network is composed of deconvolution and unpooling layers, which identify pixel-wise class labels and predict segmentation masks.
We apply the trained network to each proposal in an input image, and construct the final semantic segmentation map by combining the results from all proposals in a simple manner.
The proposed algorithm mitigates the limitations of the existing methods based on fully convolutional networks by integrating deep deconvolution network and proposal-wise prediction; our segmentation method typically identifies detailed structures and handles objects in multiple scales naturally.
Our network demonstrates outstanding performance in PASCAL VOC 2012 dataset, and we achieve the best accuracy (72.5\%) among the methods trained with no external data through ensemble with the fully convolutional network.

\end{abstract}


\section{Introduction}
\label{sec:introduction}
Convolutional neural networks (CNN) have shown excellent performance in various visual recognition problems such as image classification~\cite{Alexnet,Vgg16,Googlenet}, object detection~\cite{Rcnn,Sds}, semantic segmentation~\cite{Farabet,Zoomout}, and action recognition~\cite{JiTPAMI13,SimonyanNIPS2014}.
The representation power of CNNs leads to successful results; a combination of feature descriptors extracted from CNNs and simple off-the-shelf classifiers works very well in practice.
Encouraged by the success in classification problems, researchers start to apply CNNs to structured prediction problems, \ie, semantic segmentation~\cite{Fcn,Deeplabcrf}, human pose estimation~\cite{LiACCV14}, and so on.

Recent semantic segmentation algorithms are often formulated to solve structured pixel-wise labeling problems based on CNN~\cite{Deeplabcrf,Fcn}. 
They convert an existing CNN architecture constructed for classification to a fully convolutional network (FCN).
They obtain a coarse label map from the network by classifying every local region in image, and perform a simple deconvolution, which is implemented as bilinear interpolation, for pixel-level labeling.
Conditional random field (CRF) is optionally applied to the output map for fine segmentation~\cite{Fullycrf}.
The main advantage of the methods based on FCN is that the network accepts a whole image as an input and performs fast and accurate inference.

%
\begin{figure}[t]
\centering
\subfigure[ Inconsistent labels due to large object size ]{
\includegraphics[width=0.9\linewidth] {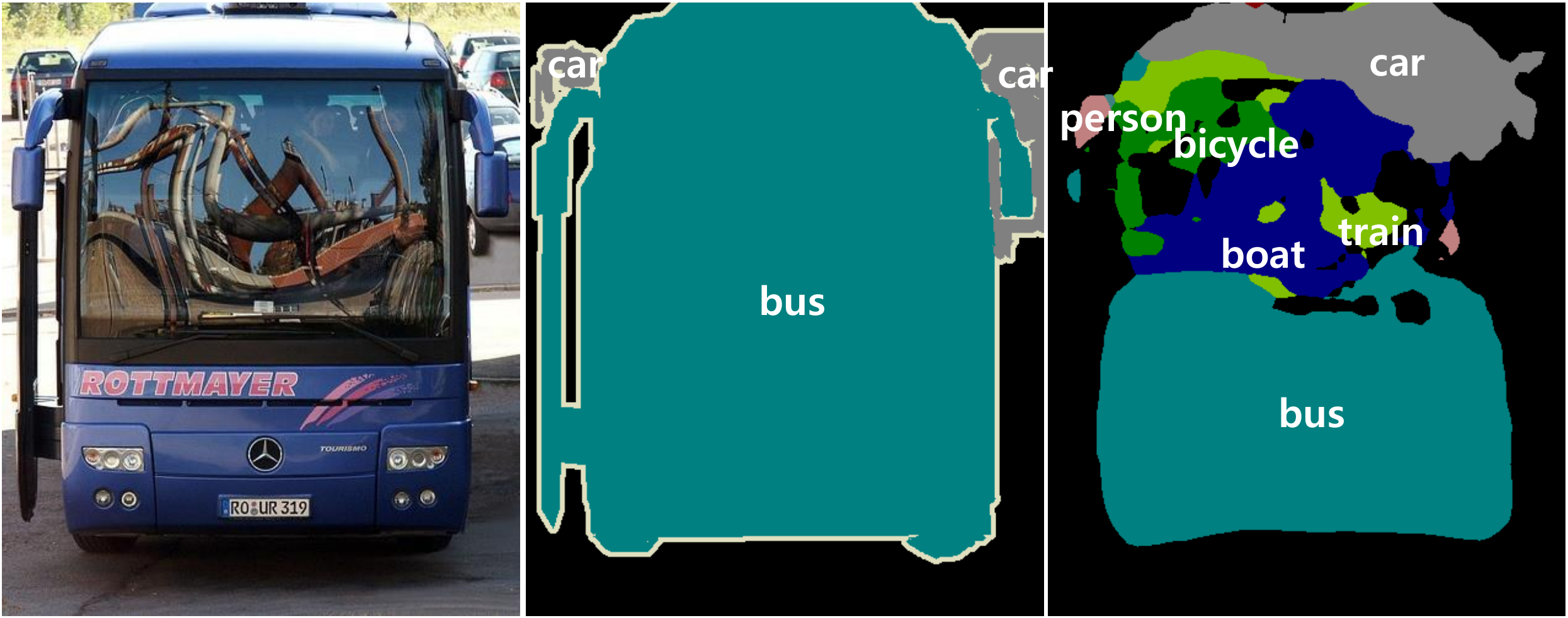}
\label{fig:fcn_limitation_a}
} \\ \vspace{-0.2cm}
\subfigure[ Missing labels due to small object size ]{
\includegraphics[width=0.9\linewidth] {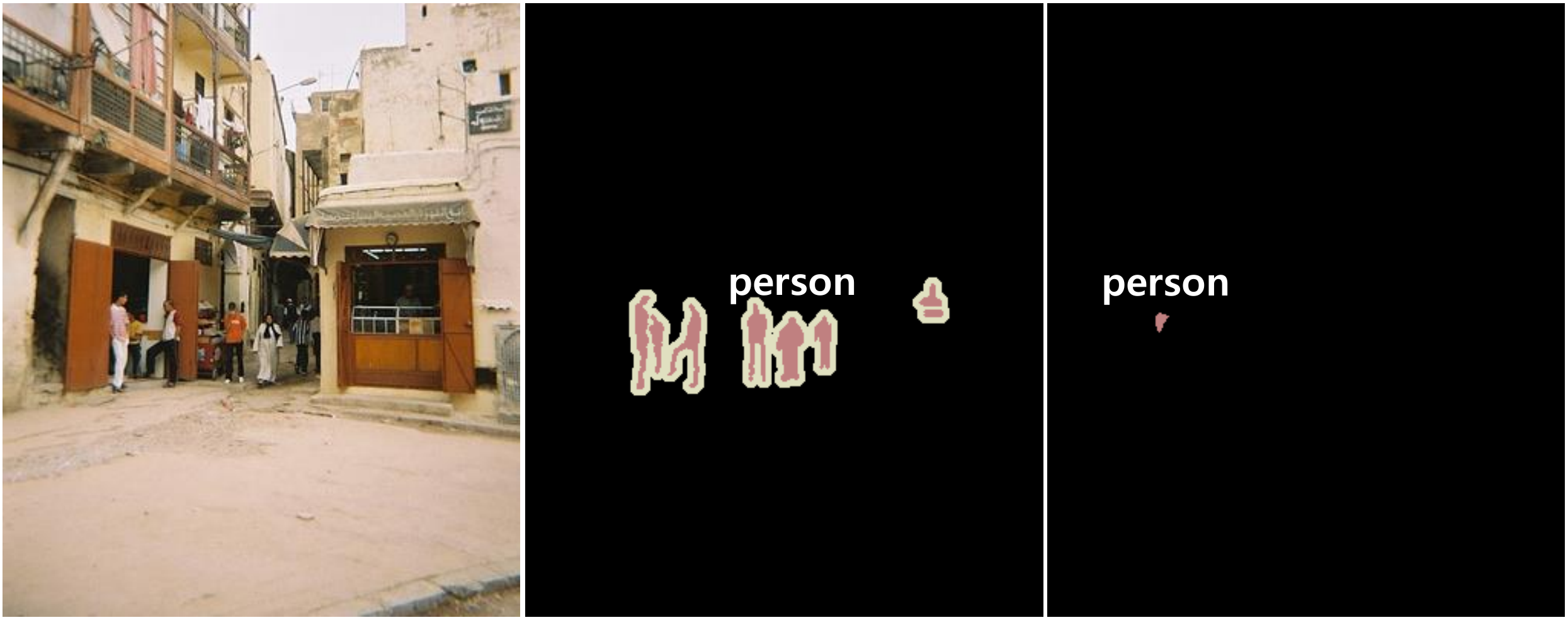}
\label{fig:fcn_limitation_b}
}
\caption{Limitations of semantic segmentation algorithms based on fully convolutional network. (Left) original image. (Center) ground-truth annotation. (Right) segmentations by \cite{Fcn} }
\label{fig:fcn_limitation}
\end{figure}
Semantic segmentation based on FCNs~\cite{Deeplabcrf,Fcn} have a couple of critical limitations.
First, the network can handle only a single scale semantics within image due to the fixed-size receptive field.
Therefore, the object that is substantially larger or smaller than the receptive field may be fragmented or mislabeled.
In other words, label prediction is done with only local information for large objects and the pixels that belong to the same object may have inconsistent labels as shown in Figure~\ref{fig:fcn_limitation_a}.
Also, small objects are often ignored and classified as background, which is illustrated in Figure~\ref{fig:fcn_limitation_b}.
Although \cite{Fcn} attempts to sidestep this limitation using skip architecture, this is not a fundamental solution and performance gain is not significant.
Second, the detailed structures of an object are often lost or smoothed because the label map, input to the deconvolutional layer, is too coarse and deconvolution procedure is overly simple.
Note that, in the original FCN~\cite{Fcn}, the label map is only $16 \times 16$ in size and is deconvolved to generate segmentation result in the original input size through bilinear interpolation.
The absence of real deconvolution in \cite{Deeplabcrf,Fcn} makes it difficult to achieve good performance.
However, recent methods ameliorate this problem using CRF~\cite{Fullycrf}.

To overcome such limitations, we employ a completely different strategy to perform semantic segmentation based on CNN.
Our main contributions are summarized below:
\begin{itemize}
\item We learn a multi-layer deconvolution network, which is composed of deconvolution, unpooling, and rectified linear unit (ReLU) layers.
Learning deconvolution network for semantic segmentation is meaningful but no one has attempted to do it yet to our knowledge.
\item The trained network is applied to individual object proposals to obtain instance-wise segmentations, which are combined for the final semantic segmentation; it is free from scale issues found in FCN-based methods and identifies finer details of an object.
\item We achieve outstanding performance using the deconvolution network trained only on PASCAL VOC 2012 dataset, and obtain the best accuracy through the ensemble with \cite{Fcn} by exploiting the heterogeneous and complementary characteristics of our algorithm with respect to FCN-based methods.
\end{itemize}
We believe that all of these three contributions help achieve the state-of-the-art performance in PASCAL VOC 2012 benchmark.

The rest of this paper is organized as follows. 
We first review related work in Section~\ref{sec:related} and describe the architecture of our network in Section~\ref{sec:supervised}.
The detailed procedure to learn a supervised deconvolution network is discussed in Section~\ref{sec:training}. 
Section~\ref{sec:image} presents how to utilize the learned deconvolution network for semantic segmentation. 
Experimental results are demonstrated in Section~\ref{sec:experiments}.


\section{Related Work}
\label{sec:related}
CNNs are very popular in many visual recognition problems and have also been applied to semantic segmentation actively.
We first summarize the existing algorithms based on supervised learning for semantic segmentation.

There are several semantic segmentation methods based on classification.
Mostajabi \etal~\cite{Zoomout} and Farabet \etal~\cite{Farabet} classify multi-scale superpixels into predefined categories and combine the classification results for pixel-wise labeling.
Some algorithms~\cite{DAICVPR15,Sds, Hypercolumns} classify region proposals and refine the labels in the image-level segmentation map to obtain the final segmentation.

Fully convolutional network (FCN)~\cite{Fcn} has driven recent breakthrough on deep learning based semantic segmentation.
In this approach, fully connected layers in the standard CNNs are interpreted as convolutions with large receptive fields, and segmentation is achieved using coarse class score maps obtained by feedforwarding an input image.
An interesting idea in this work is that a simple interpolation filter is employed for deconvolution and only the CNN part of the network is fine-tuned to learn deconvolution indirectly.
Surprisingly, the output network illustrates impressive performance on the PASCAL VOC benchmark.
Chen \etal~\cite{Deeplabcrf} obtain denser score maps within the FCN framework to predict pixel-wise labels and refine the label map using the fully connected CRF~\cite{Fullycrf}.

In addition to the methods based on supervised learning, several semantic segmentation techniques in weakly supervised settings have been proposed.
When only bounding box annotations are given for input images, \cite{Boxsup,Weaklyandsemi} refine the annotations through iterative procedures and obtain accurate segmentation outputs. On the other hand, \cite{PinheiroArxiv2015} performs semantic segmentation based only on image-level annotations in a multiple instance learning framework.

Semantic segmentation involves deconvolution conceptually, but learning deconvolution network is not very common.
Deconvolution network is introduced in \cite{Deconvnet} to reconstruct input images.
As the reconstruction of an input image is non-trivial due to max pooling layers, it proposes the unpooling operation by storing the pooled location.
Using the deconvoluton network, the input image can be reconstructed from its feature representation.
This approach is also employed to visualize activated features in a trained CNN~\cite{Visandund} and update network architecture for performance enhancement.
This visualization is useful for understanding the behavior of a trained CNN model.

\section{System Architecture}
\label{sec:supervised}

\begin{figure*}
\centering
\includegraphics[width=1\linewidth] {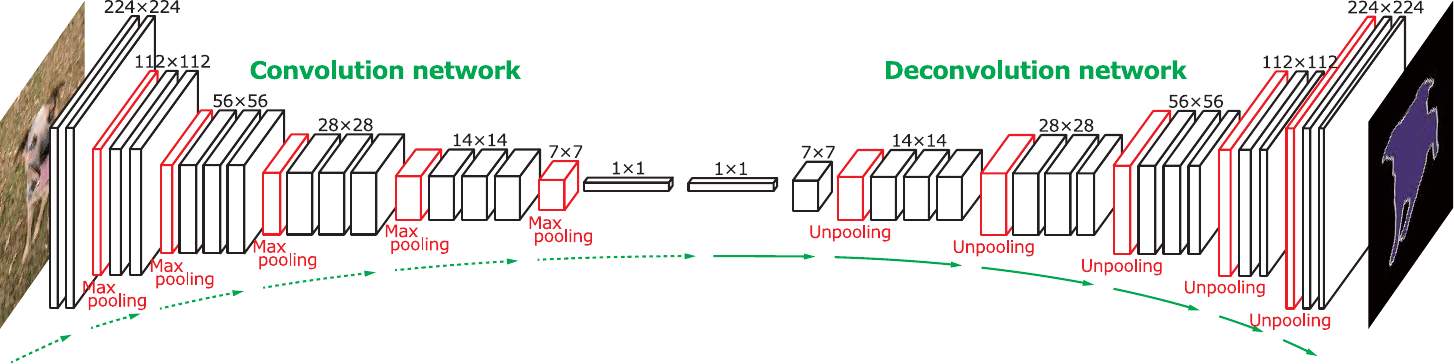}
\caption{Overall architecture of the proposed network. 
On top of the convolution network based on VGG 16-layer net, we put a multi-layer deconvolution network to generate the accurate segmentation map of an input proposal.
Given a feature representation obtained from the convolution network, dense pixel-wise class prediction map is constructed through multiple series of unpooling, deconvolution and rectification operations.
}
\label{fig:overall}
\end{figure*}

This section discusses the architecture of our deconvolution network, and describes the overall semantic segmentation algorithm.

\subsection{Architecture}

\ifdefined\paratitle {\color{blue} [Intuitive description about our network---role of convolution and deconvolution network]} \\ \fi
Figure~\ref{fig:overall} illustrates the detailed configuration of the entire deep network.
Our trained network is composed of two parts---convolution and deconvolution networks. 
The convolution network corresponds to feature extractor that transforms the input image to multidimensional feature representation, whereas the deconvolution network is a shape generator that produces object segmentation from the feature extracted from the convolution network.
The final output of the network is a probability map in the same size to input image, indicating probability of each pixel that belongs to one of the predefined classes.

\ifdefined\paratitle {\color{blue} [Detailed configuration of convolution and deconvolution network]} \\ \fi
We employ VGG 16-layer net~\cite{Vgg16} for convolutional part with its last classification layer removed. 
Our convolution network has 13 convolutional layers altogether, rectification and pooling operations are sometimes performed between convolutions, and 2 fully connected layers are augmented at the end to impose class-specific projection.
Our deconvolution network is a mirrored version of the convolution network, and has multiple series of unpooing, deconvolution, and rectification layers.
Contrary to convolution network that reduces the size of activations through feedforwarding, deconvolution network enlarges the activations through the combination of unpooling and deconvolution operations.
More details of the proposed deconvolution network is described in the following subsections.

\subsection{Deconvolution Network for Segmentation}
\label{sub:deconvolution}
We now discuss two main operations, unpooling and deconvolution, in our deconvolution network in details.

\subsubsection{Unpooling}
\label{sec:unpooling}

\ifdefined\paratitle {\color{blue} [Disadvantage of pooling in semantic segmentation]} \\ \fi
Pooling in convolution network is designed to filter noisy activations in a lower layer by abstracting activations in a receptive field with a single representative value.
Although it helps classification by retaining only robust activations in upper layers, spatial information within a receptive field is lost during pooling, which may be critical for precise localization that is required for semantic segmentation.  

\begin{figure}
\centering
\includegraphics[width=1\linewidth] {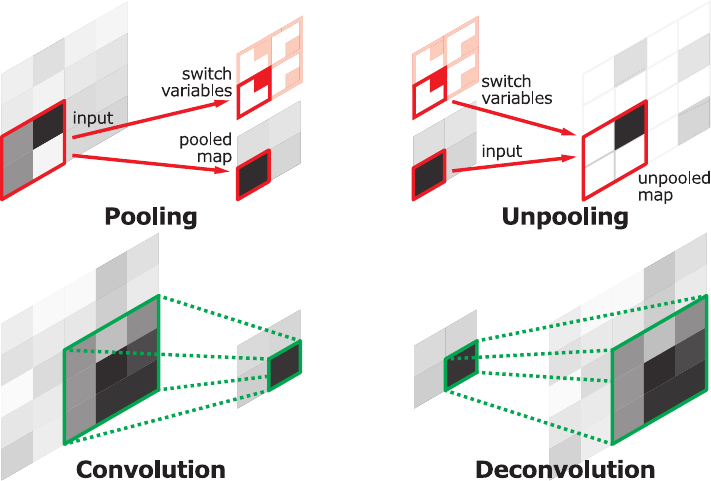}
\caption{Illustration of deconvolution and unpooling operations.}
\label{fig:deconvolution}
\end{figure}

\ifdefined\paratitle {\color{blue} [Introduction of unpooling operation]} \\ \fi
To resolve such issue, we employ unpooling layers in deconvolution network, which perform the reverse operation of pooling and reconstruct the original size of activations as illustrated in Figure~\ref{fig:deconvolution}.
To implement the unpooling operation, we follow the similar approach proposed in~\cite{Visandund,Deconvnet}. 
It records the locations of maximum activations selected during pooling operation in switch variables, which are employed to place each activation back to its original pooled location. 
This unpooling strategy is particularly useful to reconstruct the structure of input object as described in \cite{Visandund}.

%
%
\begin{figure*}[!t]
\centering
\subfigure[]{
\includegraphics[width=0.185\linewidth] {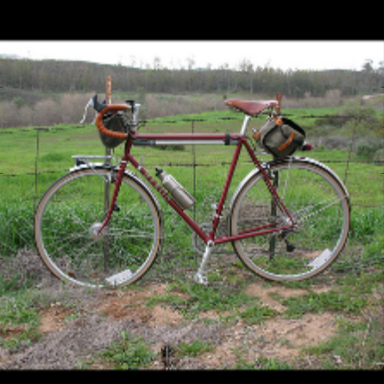}
}
\subfigure[]{
\includegraphics[width=0.185\linewidth] {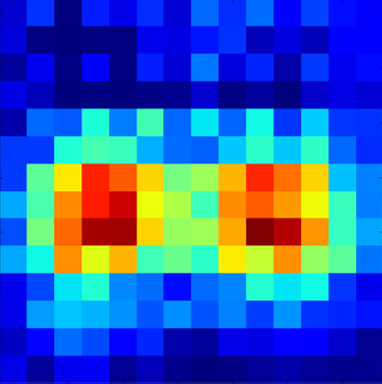}
}
\subfigure[]{
\includegraphics[width=0.185\linewidth] {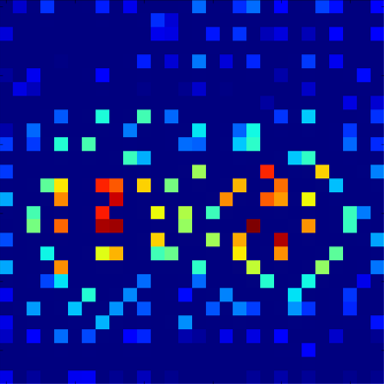}
}
\subfigure[]{
\includegraphics[width=0.185\linewidth] {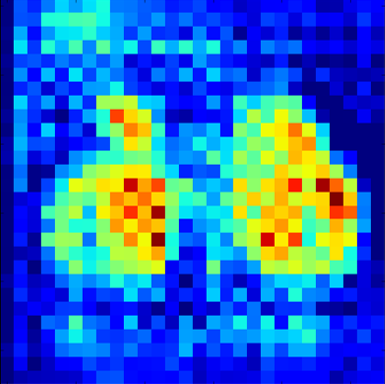}
}
\subfigure[]{
\includegraphics[width=0.185\linewidth] {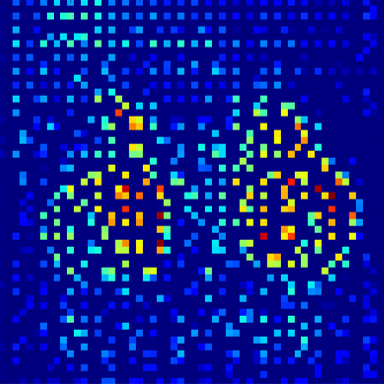} 
} \\ \vspace{-0.2cm}
\subfigure[]{
\includegraphics[width=0.185\linewidth] {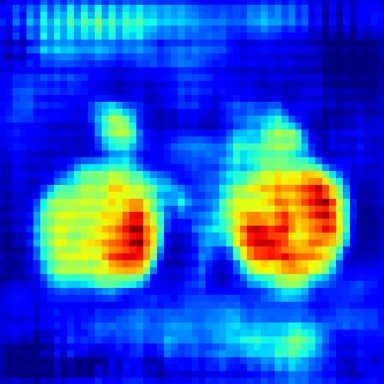}
}
\subfigure[]{
\includegraphics[width=0.185\linewidth] {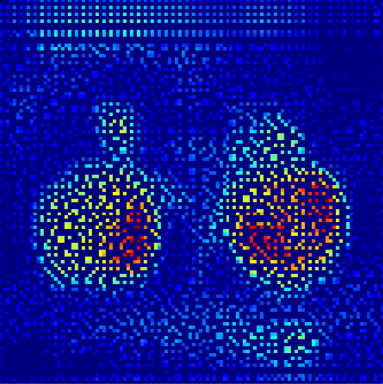}
}
\subfigure[]{
\includegraphics[width=0.185\linewidth] {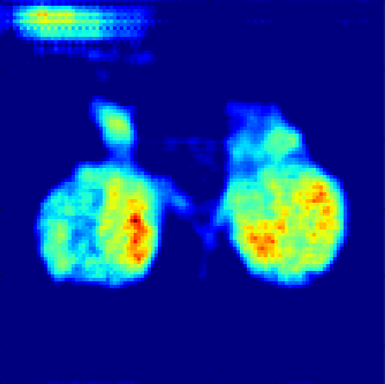}
}
\subfigure[]{
\includegraphics[width=0.185\linewidth] {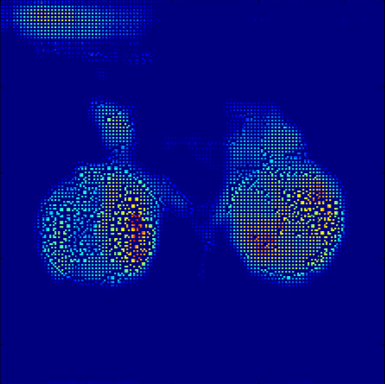}
}
\subfigure[]{
\includegraphics[width=0.185\linewidth] {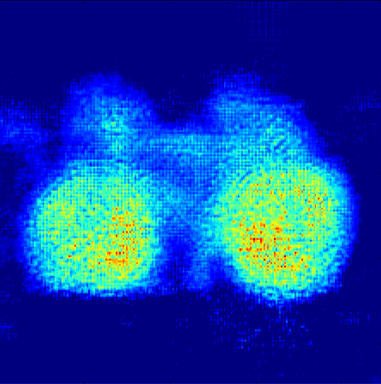}
} 
\caption{Visualization of activations in our deconvolution network. 
The activation maps from top left to bottom right correspond to the output maps from lower to higher layers in the deconvolution network.
We select the most representative activation in each layer for effective visualization.
The image in (a) is an input, and the rest are the outputs from (b) the last $14 \times 14$ deconvolutional layer, (c) the $28 \times 28$ unpooling layer, (d) the last $28 \times 28$ deconvolutional layer, (e) the $56 \times 56$ unpooling layer, (f) the last $56 \times 56$ deconvolutional layer, (g) the $112 \times 112$ unpooling layer, (h) the last $112 \times 112$ deconvolutional layer, (i) the $224 \times 224$ unpooling layer and (j) the last $224 \times 224$ deconvolutional layer. 
The finer details of the object are revealed, as the features are forward-propagated through the layers in the deconvolution network.
Note that noisy activations from background are suppressed through propagation while the activations closely related to the target classes are amplified. 
It shows that the learned filters in higher deconvolutional layers tend to capture class-specific shape information.
}
\label{fig:visdeconv}
\end{figure*}

\subsubsection{Deconvolution}
\label{sec:deconv}
\ifdefined\paratitle {\color{blue} [Descriptions on deconvolution operation]} \\ \fi
The output of an unpooling layer is an enlarged, yet sparse activation map.
The deconvolution layers densify the sparse activations obtained by unpooling through convolution-like operations with multiple learned filters.
However, contrary to convolutional layers, which connect multiple input activations within a filter window to a single activation, deconvolutional layers associate a single input activation with multiple outputs, as illustrated in Figure~\ref{fig:deconvolution}.
The output of the deconvolutional layer is an enlarged {\em and} dense activation map. 
We crop the boundary of the enlarged activation map to keep the size of the output map identical to the one from the preceding unpooling layer.
%

\ifdefined\paratitle {\color{blue} [Discussion of role of deconvolution for semantic segmentation]} \\ \fi
The learned filters in deconvolutional layers correspond to bases to reconstruct shape of an input object.
Therefore, similar to the convolution network, a hierarchical structure of deconvolutional layers are used to capture different level of shape details.
The filters in lower layers tend to capture overall shape of an object while the class-specific fine-details are encoded in the filters in higher layers.
In this way, the network directly takes class-specific shape information into account for semantic segmentation, which is often ignored in other approaches based only on convolutional layers~\cite{Deeplabcrf,Fcn}.

\subsubsection{Analysis of Deconvolution Network}
%
In the proposed algorithm, the deconvolution network is a key component for precise object segmentation. 
Contrary to the simple deconvolution in \cite{Fcn} performed on coarse activation maps, our algorithm generates object segmentation masks using deep deconvolution network, where a dense pixel-wise class probability map is obtained by successive operations of unpooling, deconvolution, and rectification.

Figure~\ref{fig:visdeconv} visualizes the outputs from the network layer by layer, which is helpful to  understand internal operations of our deconvolution network.
We can observe that coarse-to-fine object structures are reconstructed through the propagation in the deconvolutional layers; lower layers tend to capture overall coarse configuration of an object (e.g. location, shape and region), while more complex patterns are discovered in higher layers.
Note that unpooling and deconvolution play different roles for the construction of segmentation masks.
Unpooling captures \emph{example-specific} structures by tracing the original locations with strong activations back to image space.  
As a result, it effectively reconstructs the detailed structure of an object in finer resolutions.
On the other hand, learned filters in deconvolutional layers tend to capture \emph{class-specific} shapes. 
Through deconvolutions, the activations closely related to the target classes are amplified while noisy activations from other regions are suppressed effectively.
By the combination of unpooling and deconvolution, our network generates accurate segmentation maps.

\begin{figure}[!t]
\centering
\includegraphics[width=0.32\linewidth] {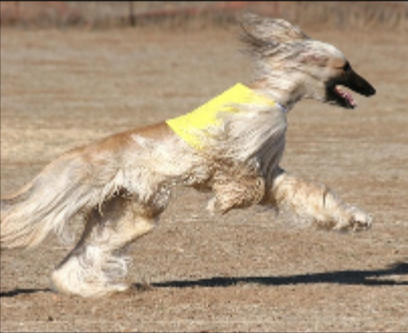} \hspace{-0.12cm}
\includegraphics[width=0.32\linewidth] {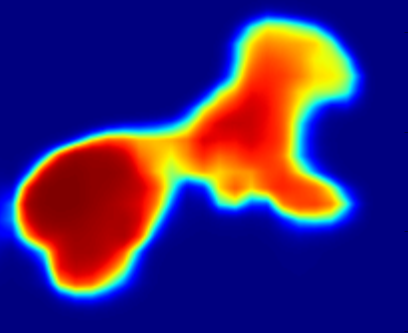} \hspace{-0.12cm}
\includegraphics[width=0.32\linewidth] {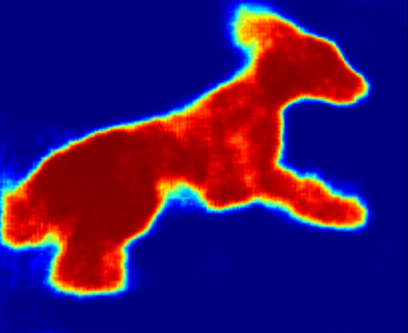} \\
\vspace{-0.15cm}
\subfigure[Input image]{
\includegraphics[width=0.32\linewidth] {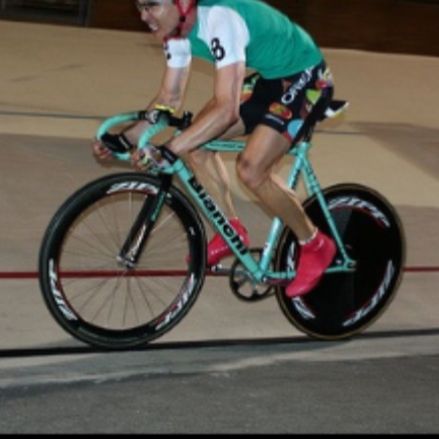}
}\hspace{-0.2cm}
\subfigure[FCN-8s]{
\includegraphics[width=0.32\linewidth] {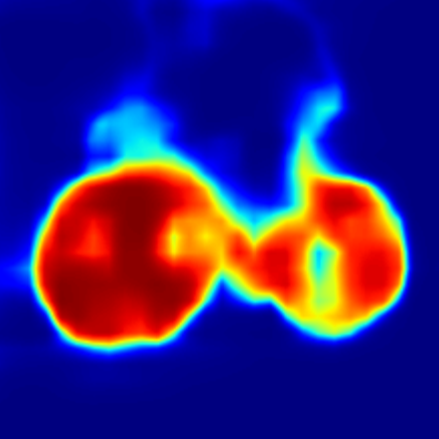}
}\hspace{-0.2cm}
\subfigure[Ours]{
\includegraphics[width=0.32\linewidth] {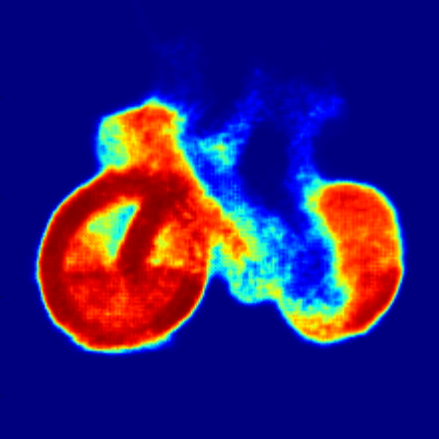}
}
\caption{
Comparison of class conditional probability maps from FCN and our network (top: dog, bottom: bicycle).
}
\label{fig:compfcn}
\end{figure}

Figure~\ref{fig:compfcn} illustrates examples of outputs from FCN-8s and the proposed network.
Compared to the coarse activation map of FCN-8s, our network constructs dense and precise activations using the deconvolution network.
%

\subsection{System Overview}
\ifdefined\paratitle {\color{blue} [our algorithm do instance-wise prediction---define what is instance-wise prediction clearly]} \\ \fi
Our algorithm poses semantic segmentation as instance-wise segmentation problem. 
That is, the network takes a sub-image potentially containing objects---which we refer to as \textit{instance(s)} afterwards---as an input and produces pixel-wise class prediction as an output. 
Given our network, semantic segmentation on a whole image is obtained by applying the network to each candidate proposals extracted from the image and aggregating outputs of all proposals to the original image space.

Instance-wise segmentation has a few advantages over image-level prediction.
It handles objects in various scales effectively and identifies fine details of objects while the approaches with fixed-size receptive fields have troubles with these issues.
Also, it alleviates training complexity by reducing search space for prediction and reduces memory requirement for training.


\section{Training}
\label{sec:training}
The entire network described in the previous section is very deep (twice deeper than \cite{Vgg16}) and contains a lot of associated parameters.
In addition, the number of training examples for semantic segmentation is relatively small compared to the size of the network---12031 PASCAL training and validation images in total.
Training a deep network with a limited number of examples is not trivial and we train the network successfully using the following ideas.

\subsection{Batch Normalization}

\ifdefined\paratitle {\color{blue} [Why training a deep neural network is difficult]} \\ \fi
It is well-known that a deep neural network is very hard to optimize due to the internal-covariate-shift problem~\cite{LOFFEARXIV15}; input distributions in each layer change over iteration during training as the parameters of its previous layers are updated.
This is problematic in optimizing very deep networks since the changes in distribution are amplified through propagation across layers.


\ifdefined\paratitle {\color{blue} [How we resolve this issue]} \\ \fi
We perform the batch normalization~\cite{LOFFEARXIV15} to reduce the internal-covariate-shift by normalizing input distributions of every layer to the standard Gaussian distribution.
For the purpose, a batch normalization layer is added to the output of every convolutional and deconvolutional layer.
We observe that the batch normalization is critical to optimize our network; it ends up with a poor local optimum without batch normalization.

\subsection{Two-stage Training}
\ifdefined\paratitle {\color{blue} [Why the training the segmentation network is still difficult---space of possible segmentation is very large]} \\ \fi
Although batch normalization helps escape local optima, the space of semantic segmentation is still very large compared to the number of training examples and the benefit to use a deconvolution network for instance-wise segmentation would be cancelled. 
Then, we employ a two-stage training method to address this issue, where we train the network with easy examples first and fine-tune the trained network with more challenging examples later.

\ifdefined\paratitle {\color{blue} [First stage training]} \\ \fi
To construct training examples for the first stage training, we crop object instances using ground-truth annotations so that an object is centered at the cropped bounding box.
By limiting the variations in object location and size, we reduce search space for semantic segmentation significantly and train the network with much less training examples successfully.
\ifdefined\paratitle {\color{blue} [Second stage training]} \\ \fi
In the second stage, we utilize object proposals to construct more challenging examples.
Specifically, candidate proposals sufficiently overlapped with ground-truth segmentations are selected for training. 
Using the proposals to construct training data makes the network more robust to the misalignment of proposals in testing, but makes training more challenging since the location and scale of an object may be significantly different across training examples.

\section{Inference}
\label{sec:image}

The proposed network is trained to perform semantic segmentation for individual instances.
Given an input image, we first generate a sufficient number of candidate proposals, and apply the trained network to obtain semantic segmentation maps of individual proposals. 
Then we aggregate the outputs of all proposals to produce semantic segmentation on a whole image.
Optionally, we take ensemble of our method with FCN~\cite{Fcn} to further improve performance. 
We describe detailed procedure in the following.

\subsection{Aggregating Instance-wise Segmentation Maps}

\ifdefined\paratitle {\color{blue} [How we aggregate outputs of individual proposals]} \\ \fi
Since some proposals may result in incorrect predictions due to misalignment to object or cluttered background, we should suppress such noises during aggregation.
The pixel-wise maximum or average of the score maps corresponding all classes turns out to be sufficiently effective to obtain robust results.

Let $g_i\in \mathrm{R}^{W\times H\times C}$ be the output score maps of the $i$th proposal, where $W\times H$ and $C$ denote the size of proposal and the number of classes, respectively.
We first put it on image space with zero padding outside $g_i$; we denote the segmentation map corresponding to $g_i$ in the original image size by $G_i$ hereafter. 
Then we construct the pixel-wise class score map of an image by aggregating the outputs of all proposals by
\begin{equation}
P(x,y,c) = \max_i G_i(x,y,c), ~~~\forall i,
\label{eq:max_agg}
\end{equation}
or
\begin{equation}
P(x,y,c) = \sum_i G_i(x,y,c), ~~~\forall i.
\label{eq:avg_agg}
\end{equation}

Class conditional probability maps in the original image space are obtained by applying softmax function to the aggregated maps obtained by Eq.~\eqref{eq:max_agg} or \eqref{eq:avg_agg}.
Finally, we apply the fully-connected CRF~\cite{Fullycrf} to the output maps for the final pixel-wise labeling, where unary potential are obtained from the pixel-wise class conditional probability maps.

\subsection{Ensemble with FCN}
\label{sec:ensemble}
\ifdefined\paratitle {\color{blue} [There are complementary characteristics of our network and FCN, and it is possible to achieve better results by combining outputs of two networks. ] } \\ \fi
Our algorithm based on the deconvolution network has complementary characteristics to the approaches relying on FCN; our deconvolution network is appropriate to capture the fine-details of an object, whereas FCN is typically good at extracting the overall shape of an object.
In addition, instance-wise prediction is useful for handling objects with various scales, while fully convolutional network with a coarse scale may be advantageous to capture context within image.
Exploiting these heterogeneous properties may lead to better results, and we take advantage of the benefit of both algorithms through ensemble.

\ifdefined\paratitle {\color{blue} [How to aggregate (technically)] } \\ \fi
We develop a simple method to combine the outputs of both algorithms.
Given two sets of class conditional probability maps of an input image computed independently by the proposed method and FCN, we compute the mean of both output maps and apply the CRF to obtain the final semantic segmentation.

\section{Experiments}
\label{sec:experiments}
This section first describes our implementation details and experiment setup.
Then, we analyze and evaluate the proposed network in various aspects.

\begin{table*}[!t] \footnotesize
\centering
\caption{Evaluation results on PASCAL VOC 2012 test set. (Asterisk ($*$) denotes the algorithms  trained with additional data.)} \vspace{0.1cm}
\begin{tabular}
{
@{}C{2.7cm}@{}|@{}C{0.68cm}@{}C{0.66cm}@{}C{0.66cm}@{}C{0.66cm}@{}C{0.66cm}@{}C{0.66cm}@{}C{0.66cm}@{}C{0.66cm}@{}C{0.66cm}@{}C{0.66cm}@{}C{0.66cm}@{}C{0.66cm}@{}C{0.66cm}@{}C{0.66cm}@{}C{0.75cm}@{}C{0.75cm}@{}C{0.66cm}@{}C{0.66cm}@{}C{0.66cm}@{}C{0.66cm}@{}C{0.66cm}@{}|@{}C{0.75cm}@{}
}
\hline
Method&bkg&areo&bike&bird&boat&bottle&bus&car&cat&chair&cow&table&dog&horse&mbk&person&plant&sheep&sofa&train&tv&mean\\
\hline
Hypercolumn~\cite{Hypercolumns}&88.9&68.4&27.2&68.2&47.6&61.7&76.9&72.1&71.1&24.3&59.3&44.8&62.7&59.4&73.5&70.6&52.0&63.0&38.1&60.0&54.1&59.2\\
MSRA-CFM~\cite{DAICVPR15}&87.7&75.7&26.7&69.5&48.8&65.6&81.0&69.2&73.3&30.0&68.7&51.5&69.1&68.1&71.7&67.5&50.4&66.5&44.4&58.9&53.5&61.8\\
FCN8s~\cite{Fcn}&91.2&76.8&34.2&68.9&49.4&60.3&75.3&74.7&77.6&21.4&62.5&46.8&71.8&63.9&76.5&73.9&45.2&72.4&37.4&70.9&55.1&62.2\\
TTI-Zoomout-16~\cite{Zoomout}&89.8&81.9&35.1&78.2&57.4&56.5&80.5&74.0&79.8&22.4&69.6&53.7&74.0&76.0&76.6&68.8&44.3&70.2&40.2&68.9&55.3&64.4\\
DeepLab-CRF~\cite{Deeplabcrf}&93.1&84.4&\bf{54.5}&81.5&63.6&65.9&85.1&79.1&83.4&30.7&74.1&59.8&79.0&76.1&83.2&80.8&\bf{59.7}&82.2&50.4&73.1&63.7&71.6\\
\hline

DeconvNet &92.7&85.9&42.6&78.9&62.5&66.6&87.4&77.8&79.5&26.3&73.4&60.2&70.8&76.5&79.6&77.7&58.2&77.4&52.9&75.2&59.8&69.6\\


DeconvNet+CRF &92.9&87.8&41.9&80.6&63.9&67.3&\bf{88.1}&78.4&81.3&25.9&73.7&61.2&72.0&77.0&79.9&78.7&59.5&78.3&\bf{55.0}&75.2&61.5&70.5\\

EDeconvNet &92.9&88.4&39.7&79.0&63.0&67.7&87.1&\bf{81.5}&84.4&27.8&76.1&61.2&78.0&79.3&83.1&79.3&58.0&82.5&52.3&80.1&64.0&71.7\\

EDeconvNet+CRF &93.1&{\bf 89.9}&39.3&79.7&63.9&\bf{68.2}&87.4&81.2&\bf{86.1}&28.5&\bf{77.0}&62.0&79.0&\bf{80.3}&\bf{83.6}&80.2&58.8&\bf{83.4}&54.3&\bf{80.7}&65.0&\bf{72.5}\\
\hline
* WSSL~\cite{Weaklyandsemi}&93.2&85.3&36.2&\bf{84.8}&61.2&67.5&84.7&81.4&81.0&\bf{30.8}&73.8&53.8&77.5&76.5&82.3&\bf{81.6}&56.3&78.9&52.3&76.6&63.3&70.4\\
* BoxSup~\cite{Boxsup}&\bf{93.6}&86.4&35.5&79.7&\bf{65.2}&65.2&84.3&78.5&83.7&30.5&76.2&\bf{62.6}&\bf{79.3}&76.1&82.1&81.3&57.0&78.2&\bf{55.0}&72.5&\bf{68.1}&71.0\\
\hline
\end{tabular}
\label{tab:voc_result}
\end{table*}

\subsection{Implementation Details}
\label{sec:impledetail}

\paragraph{Network Configuration}
Table~\ref{tab:architecture_tab} summarizes the detailed configuration of the proposed network  presented in Figure~\ref{fig:overall}.
Our network has symmetrical configuration of convolution and deconvolution network centered around the 2nd fully-connected layer (fc7).
The input and output layers correspond to input image and class conditional probability maps, respectively.
The network contains approximately 252M parameters in total.

\paragraph{Dataset}
We employ PASCAL VOC 2012 segmentation dataset~\cite{Pascalvoc} for training and testing the proposed deep network.
For training, we use augmented segmentation annotations from~\cite{Hariharan}, where all training and validation images are used to train our network. 
The performance of our network is evaluated on test images.
Note that only the images in PASCAL VOC 2012 datasets are used for training in our experiment, whereas some state-of-the-art algorithms~\cite{Boxsup, Weaklyandsemi} employ additional data to improve performance.

\paragraph{Training Data Construction}
We employ a two-stage training strategy and use a separate training dataset in each stage.
To construct training examples for the first stage, we draw a tight bounding box corresponding to each annotated object in training images, and extend the box 1.2 times larger to include local context around the object. 
Then we crop the window using the extended bounding box to obtain a training example.
The class label for each cropped region is provided based only on the object located at the center while all other pixels are labeled as background. 
In the second stage, each training example is extracted from object proposal~\cite{Edgebox}, where
all relevant class labels are used for annotation.
We employ the same post-processing as the one used in the first stage to include context.
For both datasets, we maintain the balance for the number of examples across classes by adding redundant examples for the classes with limited number of examples.
To augment training data, we transform an input example to a $250\times250$ image and randomly crop the image to $224\times224$ with optional horizontal flipping in a similar way to \cite{Vgg16}.
The number of training examples is 0.2M and 2.7M in the first and the second stage, respectively, which is sufficiently large to train the deconvolution network from scratch.

\paragraph{Optmization}
We implement the proposed network based on Caffe~\cite{caffearxiv} framework. 
The standard stochastic gradient descent with momentum is employed for optimization, where initial learning rate, momentum and weight decay are set to 0.01, 0.9 and 0,0005, respectively. 
We initialize the weights in the convolution network using VGG 16-layer net pre-trained on ILSVRC~\cite{Imagenet} dataset, while the weights in the deconvolution network are initialized with zero-mean Gaussians.
We remove the drop-out layers due to batch normalization, and reduce learning rate in an order of magnitude whenever validation accuracy does not improve. 
Although our final network is learned with both train and validation datasets, learning rate adjustment based on validation accuracy still works well according to our experience.
The network converges after approximately 20K and 40K SGD iterations with mini-batch of 64 samples in the first and second stage training, respectively.
Training takes 6 days (2 days for the first stage and 4 days for the second stage) in a single Nvidia GTX Titan X GPU with 12G memory.

\paragraph{Inference}
We employ edge-box~\cite{Edgebox} to generate object proposals. 
For each testing image, we generate approximately 2000 object proposals, and select top 50 proposals based on their objectness scores. 
We observe that this number is sufficient to obtain accurate segmentation in practice. 
To obtain pixel-wise class conditional probability maps for a whole image, we compute pixel-wise maximum to aggregate proposal-wise predictions as in Eq.~\eqref{eq:max_agg}.

\begin{figure*}[!t]
\centering
\includegraphics[width=0.98\linewidth] {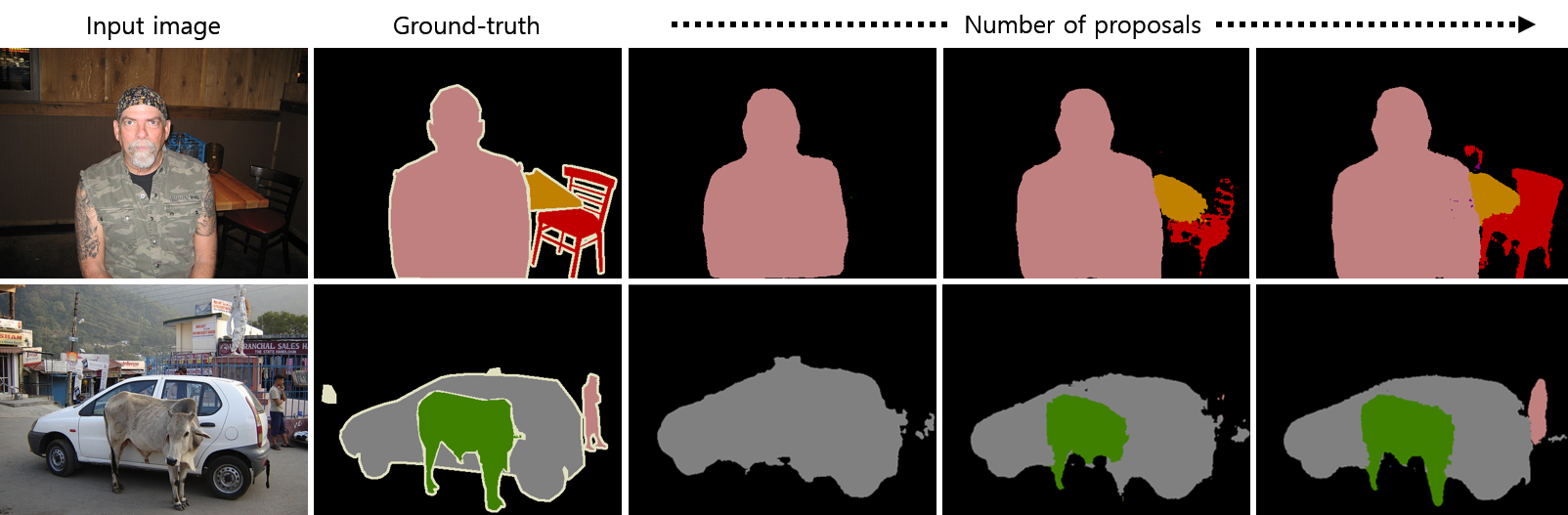}\\
\caption{
Benefit of instance-wise prediction. 
We aggregate the proposals in a decreasing order of their sizes.
The algorithm identifies finer object structures through iterations by handling multi-scale objects effectively.
}
\label{fig:instancewise}
\end{figure*}

\subsection{Evaluation on Pascal VOC}
We evaluate our network on PASCAL VOC 2012 benchmark~\cite{Pascalvoc},  which contains 1456 test images and involves 20 object categories.
We adopt {\em comp6} evaluation protocol that measures scores based on Intersection over Union (IoU) between ground truth and predicted segmentations. 

The quantitative results of the proposed algorithm and the competitors are presented in Table~\ref{tab:voc_result}\footnote{All numbers in this table are from the officially published {\em papers}, not from the leaderboard, including the ones in arXiv.}, where our method is denoted by DeconvNet.
The performance of DeconvNet is competitive to the state-of-the-art methods. 
The CRF~\cite{Fullycrf} as post-processing enhances accuracy by approximately 1\% point.
%
We further improve performance through an ensemble with FCN-8s.
It improves mean IoU about 10.3\% and 3.1\% point with respect to FCN-8s and our DeconvNet, respectively, which is notable considering relatively low accuracy of FCN-8s.
We believe that this is because our method and FCN have complementary characteristics as discussed in Section~\ref{sec:ensemble}; this property differentiates our algorithm from the existing ones based on FCN.
Our ensemble method with FCN-8s denoted by EDeconvNet achieves the best accuracy among methods trained only on PASCAL VOC data. 

Figure~\ref{fig:instancewise} demonstrates effectiveness of instance-wise prediction for accurate segmentation. 
We aggregate the proposals in a decreasing order of their sizes and observe the progress of segmentation.
As the number of aggregated proposals increases, the algorithm identifies finer object structures, which are typically captured by small proposals.

\begin{figure*}[!t]
\centering
\vspace{-0.15cm}
\includegraphics[width=0.98\linewidth] {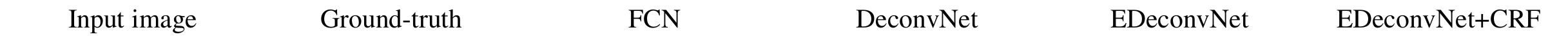}\\ \vspace{-0.2cm}
\subfigure[Examples that our method produces better results than FCN~\cite{Fcn}.]{
\begin{minipage}{1\textwidth}
\centering
\includegraphics[width=0.16\linewidth] {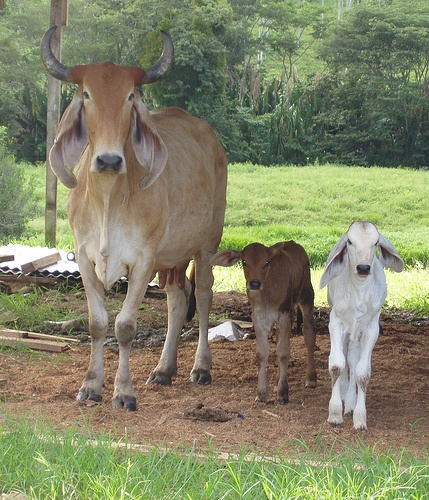}
\includegraphics[width=0.16\linewidth] {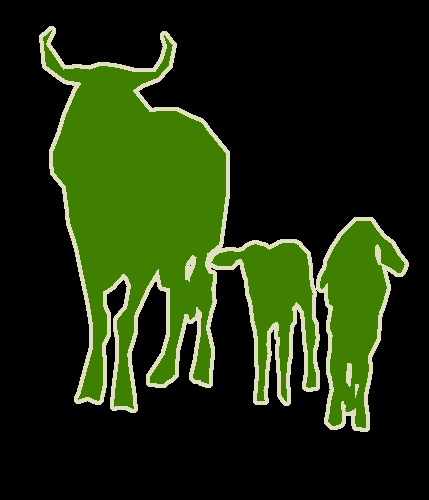}
\includegraphics[width=0.16\linewidth] {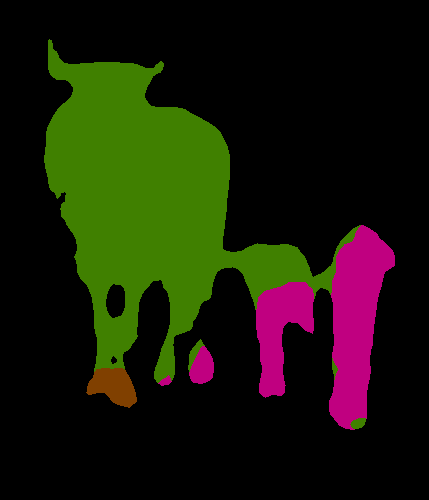}
\includegraphics[width=0.16\linewidth] {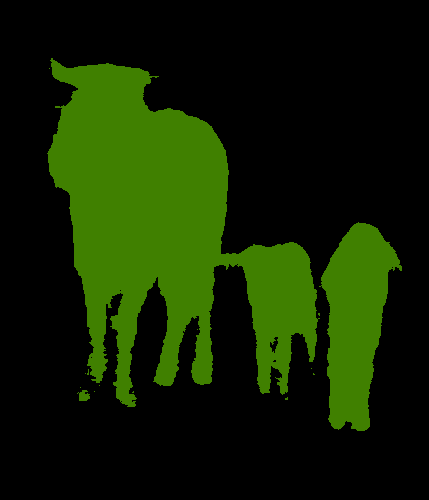}
\includegraphics[width=0.16\linewidth] {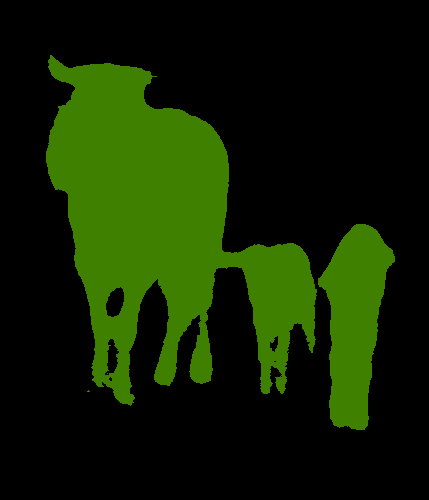} 
\includegraphics[width=0.16\linewidth] {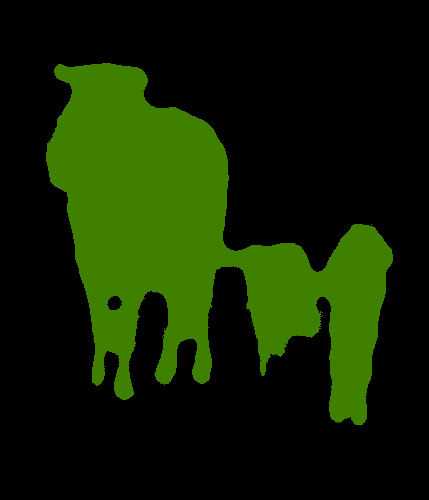} \\
\includegraphics[width=0.16\linewidth] {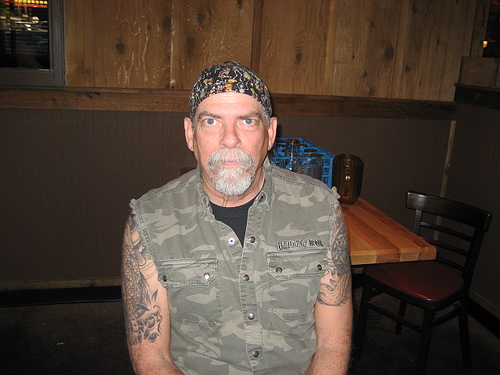}
\includegraphics[width=0.16\linewidth] {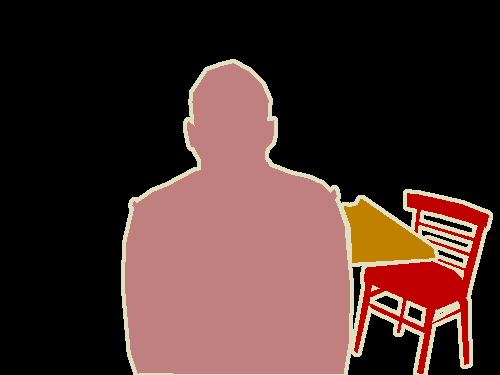}
\includegraphics[width=0.16\linewidth] {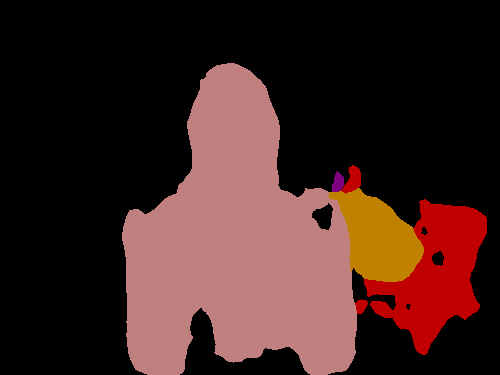}
\includegraphics[width=0.16\linewidth] {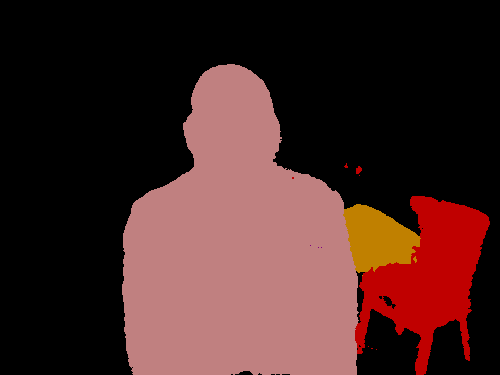}
\includegraphics[width=0.16\linewidth] {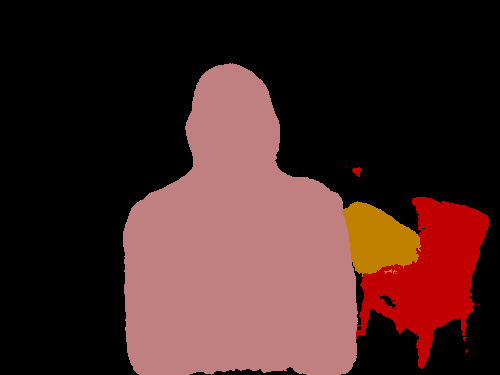} 
\includegraphics[width=0.16\linewidth] {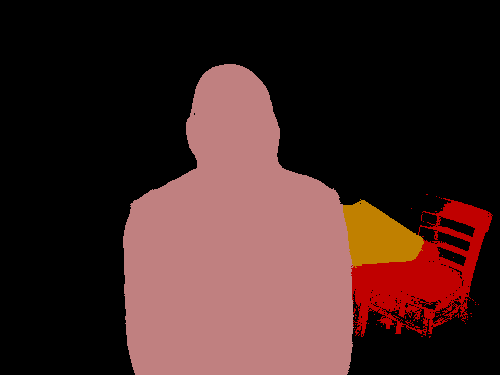} \\
\includegraphics[width=0.16\linewidth] {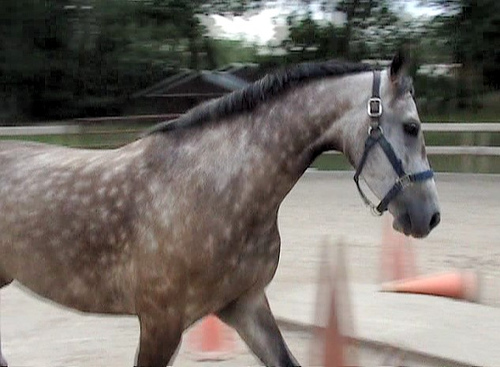}
\includegraphics[width=0.16\linewidth] {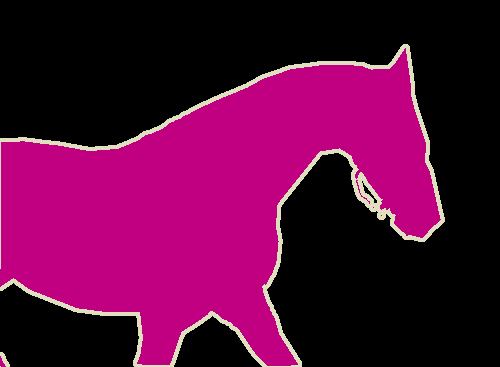}
\includegraphics[width=0.16\linewidth] {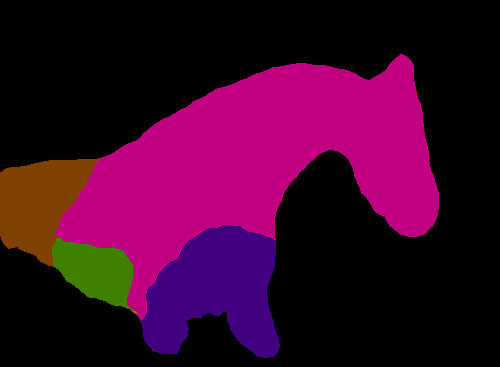}
\includegraphics[width=0.16\linewidth] {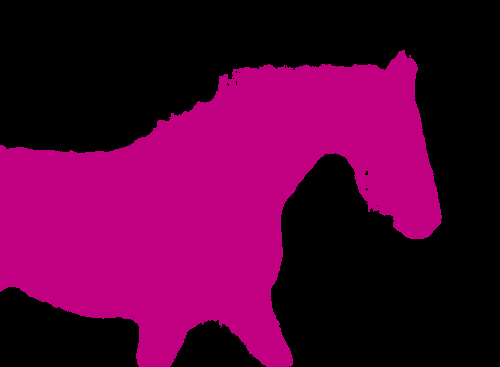}
\includegraphics[width=0.16\linewidth] {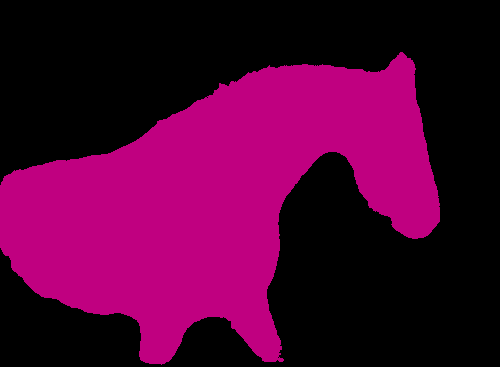} 
\includegraphics[width=0.16\linewidth] {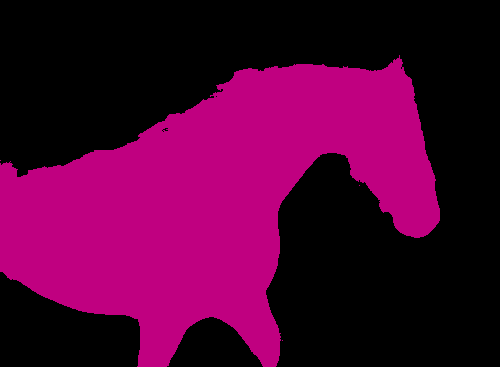} \\
\includegraphics[width=0.16\linewidth] {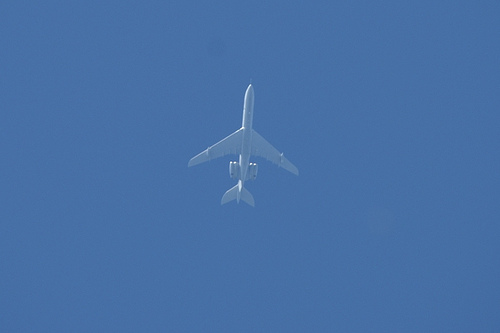}
\includegraphics[width=0.16\linewidth] {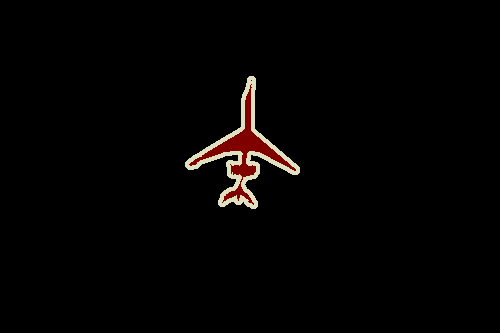}
\includegraphics[width=0.16\linewidth] {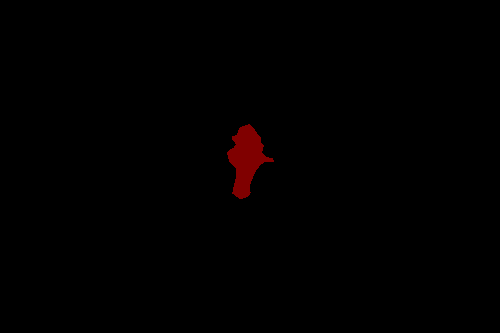}
\includegraphics[width=0.16\linewidth] {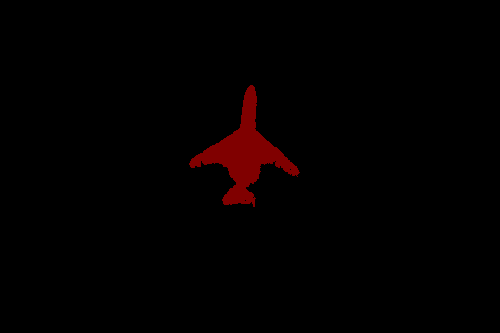}
\includegraphics[width=0.16\linewidth] {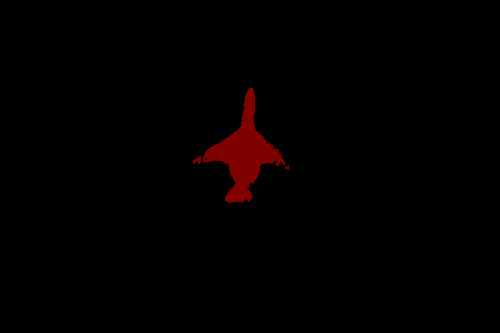}
\includegraphics[width=0.16\linewidth] {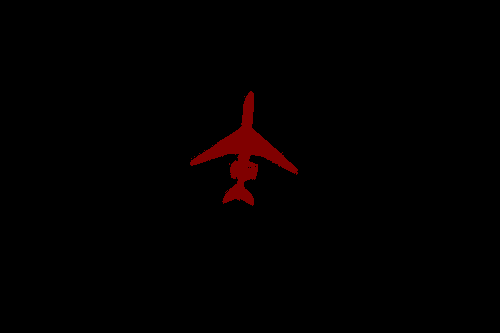} \\\vspace{0.1cm}
\end{minipage}
\label{fig:qualitative_fcn}
}\\\vspace{-0.2cm}
\subfigure[Examples that FCN produces better results than our method.]{
\begin{minipage}{1\textwidth}
\centering
\includegraphics[width=0.16\linewidth] {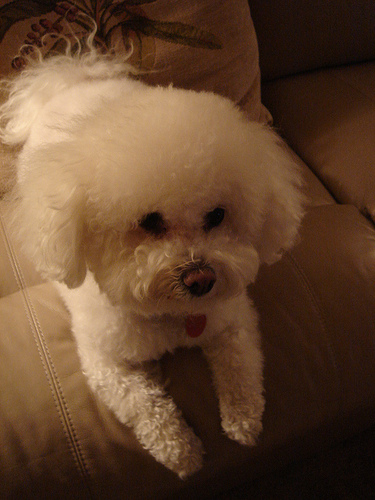}
\includegraphics[width=0.16\linewidth] {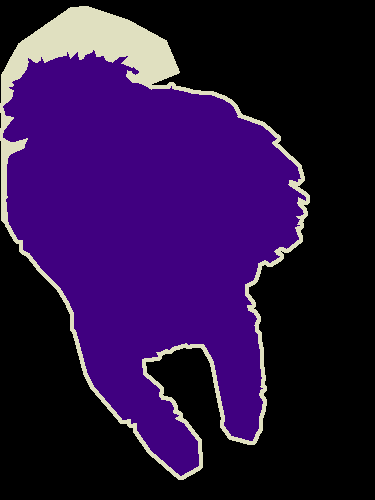}
\includegraphics[width=0.16\linewidth] {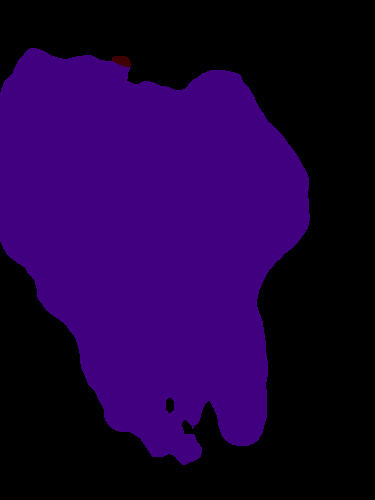}
\includegraphics[width=0.16\linewidth] {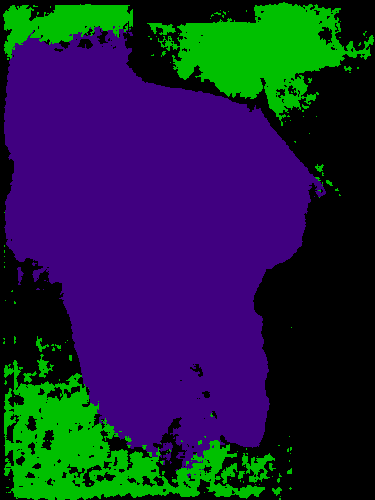}
\includegraphics[width=0.16\linewidth] {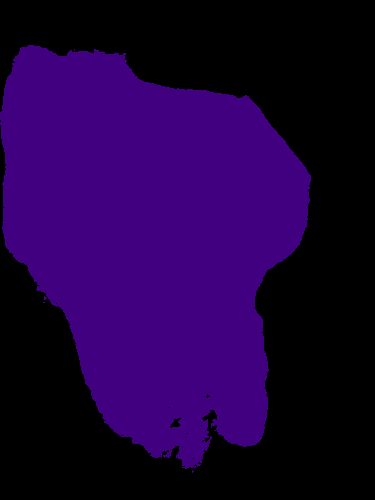}
\includegraphics[width=0.16\linewidth] {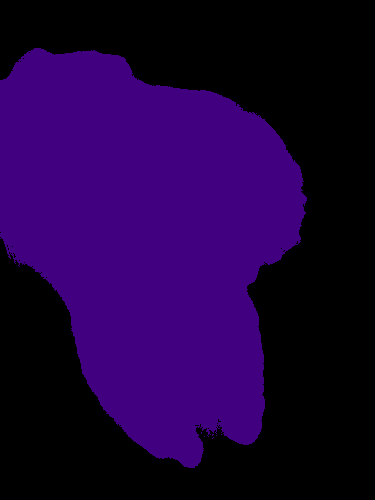} \\ 
\includegraphics[width=0.16\linewidth] {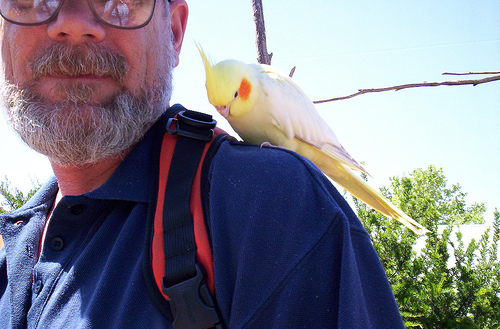}
\includegraphics[width=0.16\linewidth] {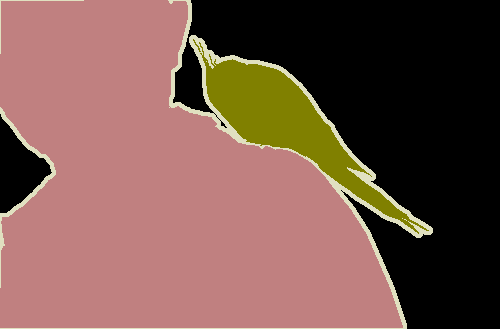}
\includegraphics[width=0.16\linewidth] {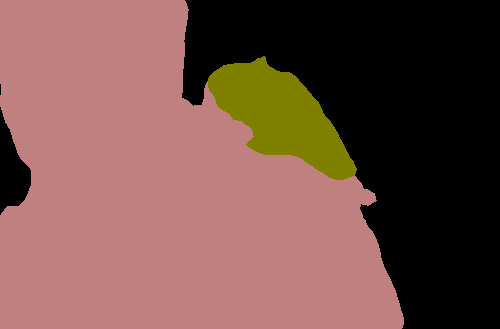}
\includegraphics[width=0.16\linewidth] {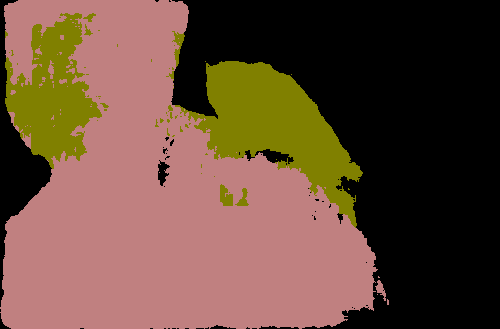}
\includegraphics[width=0.16\linewidth] {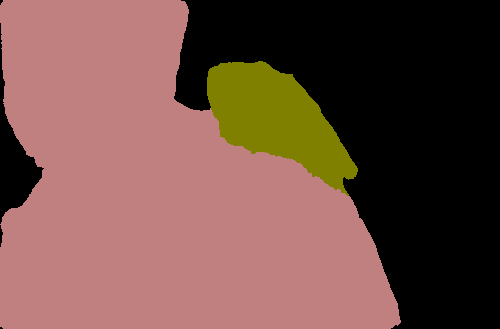}
\includegraphics[width=0.16\linewidth] {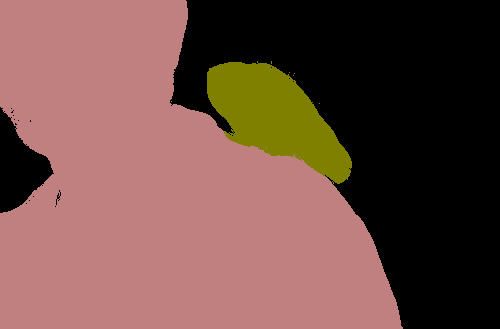} \\ \vspace{0.1cm}
\end{minipage}
\label{fig:qualitative_deconvnet}
} \\\vspace{-0.2cm}
\subfigure[Examples that inaccurate predictions from our method and FCN are improved by ensemble.]{
\begin{minipage}{1\textwidth}
\centering
\includegraphics[width=0.16\linewidth] {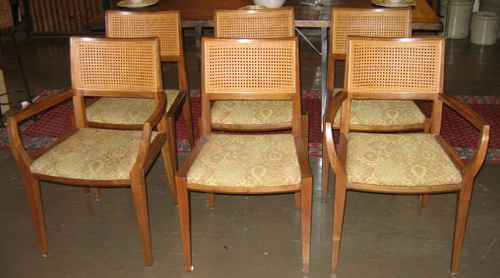}
\includegraphics[width=0.16\linewidth] {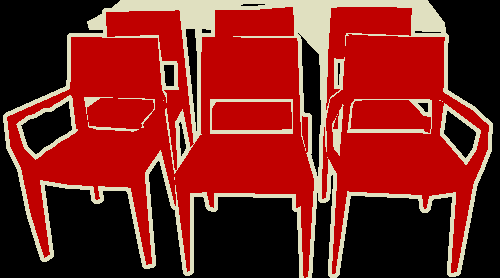}
\includegraphics[width=0.16\linewidth] {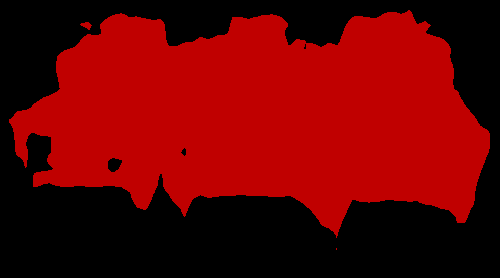}
\includegraphics[width=0.16\linewidth] {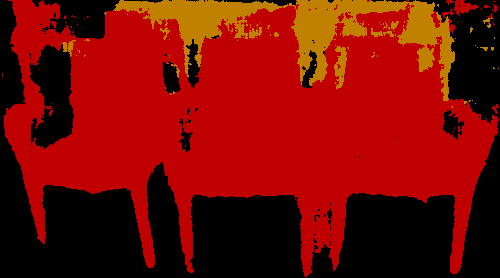}
\includegraphics[width=0.16\linewidth] {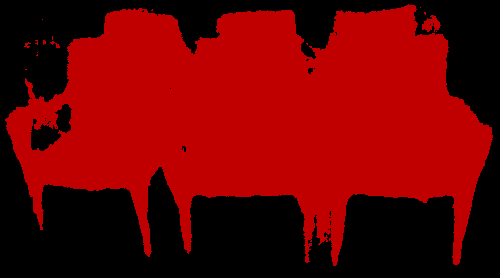}
\includegraphics[width=0.16\linewidth] {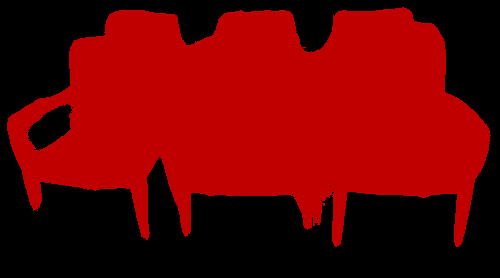}\\ 
\includegraphics[width=0.16\linewidth] {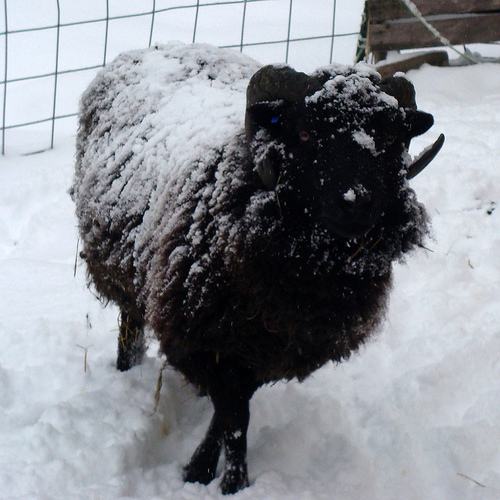}
\includegraphics[width=0.16\linewidth] {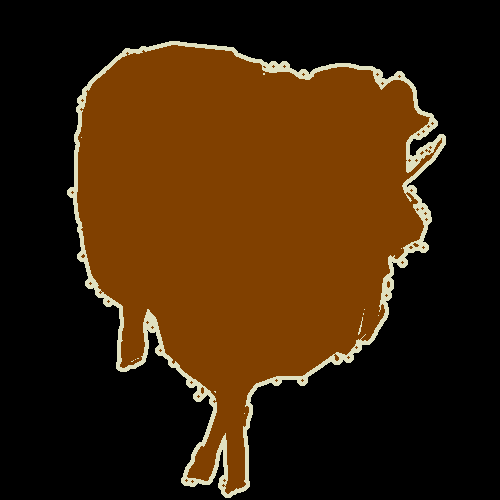}
\includegraphics[width=0.16\linewidth] {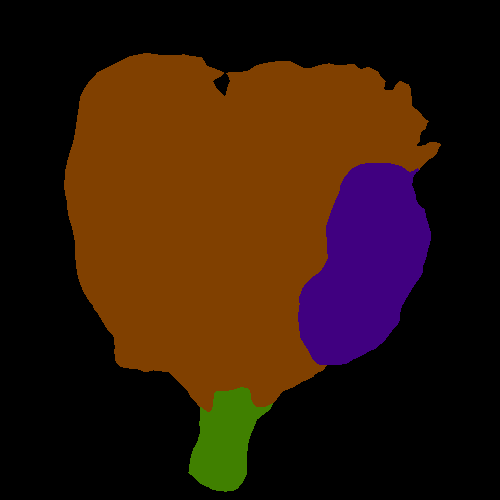}
\includegraphics[width=0.16\linewidth] {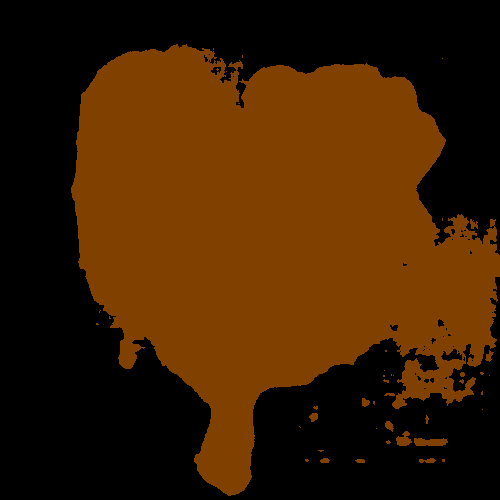}
\includegraphics[width=0.16\linewidth] {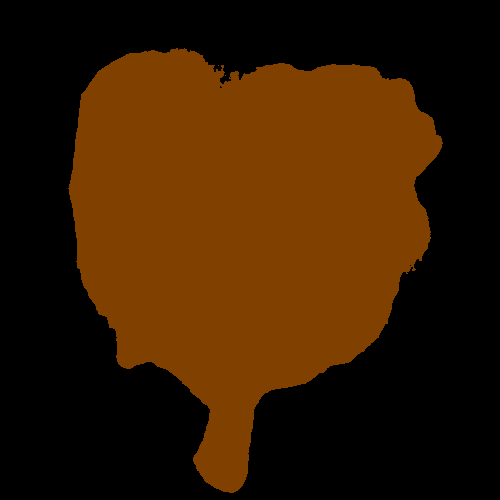}
\includegraphics[width=0.16\linewidth] {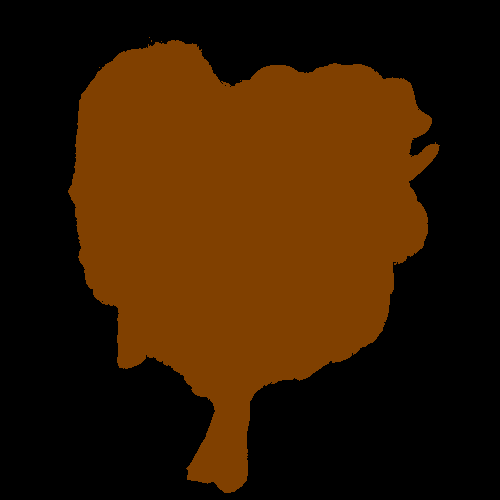}\\ \vspace{0.11cm}
\end{minipage}
\label{fig:qualitative_ensemble}
}

\caption{
Example of semantic segmentation results on PASCAL VOC 2012 validation images. 
Note that the proposed method and FCN have complementary characteristics for semantic segmentation, and the combination of both methods improves accuracy through ensemble.
Although CRF removes some noises, it does not improve quantitative performance of our algorithm significantly.
}
\label{fig:qualitative}
\end{figure*}

The qualitative results of DeconvNet, FCN and their ensemble are presented in Figure~\ref{fig:qualitative}. 
Overall, DeconvNet produces fine segmentations compared to FCN, and handles multi-scale objects effectively through instance-wise prediction.
FCN tends to fail in labeling too large or small objects (Figure~\ref{fig:qualitative_fcn}) due to its fixed-size receptive field.
Our network sometimes returns noisy predictions (Figure~\ref{fig:qualitative_deconvnet}), when the proposals are misaligned or located at background regions.
%
The ensemble with FCN-8s produces much better results as observed in Figure~\ref{fig:qualitative_fcn} and \ref{fig:qualitative_deconvnet}.
Note that inaccurate predictions from both FCN and DeconvNet are sometimes corrected by ensemble  as shown in Figure~\ref{fig:qualitative_ensemble}.
Adding CRF to ensemble improves quantitative performance, although the improvement is not significant. 

\begin{table}[!t] \footnotesize
\centering
\caption{Detailed configuration of the proposed network. ``conv'' and ``deconv'' denote layers in convolution and deconvolution network, respectively, while numbers next to each layer name mean the order of the corresponding layer in the network.
ReLU layers are omitted from the table for brevity. } 
\vspace{0.1cm}
\begin{tabular}
{
@{}C{2cm}@{}|@{}C{1.5cm}@{}C{1cm}@{}C{1cm}@{}|@{}C{2.5cm}@{}
}
\hline
name&kernel size&stride&pad&output size\\
\hline
input&-&-&-&$224\times224\times3$\\
\hline
conv1-1&$3\times3$&1&1&$224\times224\times64$\\
conv1-2&$3\times3$&1&1&$224\times224\times64$\\
\hline
pool1&$2\times2$&2&0&$112\times112\times64$\\
\hline
conv2-1&$3\times3$&1&1&$112\times112\times128$\\
conv2-2&$3\times3$&1&1&$112\times112\times128$\\
\hline
pool2&$2\times2$&2&0&$56\times56\times128$\\
\hline
conv3-1&$3\times3$&1&1&$56\times56\times256$\\
conv3-2&$3\times3$&1&1&$56\times56\times256$\\
conv3-3&$3\times3$&1&1&$56\times56\times256$\\
\hline
pool3&$2\times2$&2&0&$28\times28\times256$\\
\hline
conv4-1&$3\times3$&1&1&$28\times28\times512$\\
conv4-2&$3\times3$&1&1&$28\times28\times512$\\
conv4-3&$3\times3$&1&1&$28\times28\times512$\\
\hline
pool4&$2\times2$&2&0&$14\times14\times512$\\
\hline
conv5-1&$3\times3$&1&1&$14\times14\times512$\\
conv5-2&$3\times3$&1&1&$14\times14\times512$\\
conv5-3&$3\times3$&1&1&$14\times14\times512$\\
\hline
pool5&$2\times2$&2&0&$7\times7\times512$\\
\hline
fc6&$7\times7$&1&0&$1\times1\times4096$\\
fc7&$1\times1$&1&0&$1\times1\times4096$\\
\hline
deconv-fc6&$7\times7$&1&0&$7\times7\times512$\\
\hline
unpool5&$2\times2$&2&0&$14\times14\times512$\\
\hline
deconv5-1&$3\times3$&1&1&$14\times14\times512$\\
deconv5-2&$3\times3$&1&1&$14\times14\times512$\\
deconv5-3&$3\times3$&1&1&$14\times14\times512$\\
\hline
unpool4&$2\times2$&2&0&$28\times28\times512$\\
\hline
deconv4-1&$3\times3$&1&1&$28\times28\times512$\\
deconv4-2&$3\times3$&1&1&$28\times28\times512$\\
deconv4-3&$3\times3$&1&1&$28\times28\times256$\\
\hline
unpool3&$2\times2$&2&0&$56\times56\times256$\\
\hline
deconv3-1&$3\times3$&1&1&$56\times56\times256$\\
deconv3-2&$3\times3$&1&1&$56\times56\times256$\\
deconv3-3&$3\times3$&1&1&$56\times56\times128$\\
\hline
unpool2&$2\times2$&2&0&$112\times112\times128$\\
\hline
deconv2-1&$3\times3$&1&1&$112\times112\times128$\\
deconv2-2&$3\times3$&1&1&$112\times112\times64$\\
\hline
unpool1&$2\times2$&2&0&$224\times224\times64$\\
\hline
deconv1-1&$3\times3$&1&1&$224\times224\times64$\\
deconv1-2&$3\times3$&1&1&$224\times224\times64$\\
\hline
output&$1\times1$&1&1&$224\times224\times21$\\
\hline
\end{tabular}
\label{tab:architecture_tab}
\end{table}
\section{Conclusion}
We proposed a novel semantic segmentation algorithm by learning a deconvolution network.
The proposed deconvolution network is suitable to generate dense and precise object segmentation masks since coarse-to-fine structures of an object is reconstructed progressively through a sequence of deconvolution operations.
Our algorithm based on instance-wise prediction is advantageous to handle object scale variations by eliminating the limitation of fixed-size receptive field in the fully convolutional network.
We further proposed an ensemble approach, which combines the outputs of the proposed algorithm and FCN-based method, and achieved substantially better performance thanks to complementary characteristics of both algorithms.
Our network demonstrated the state-of-the-art performance in PASCAL VOC 2012 segmentation benchmark among the methods trained with no external data.

\newpage

{\small
\bibliographystyle{ieee}
\bibliography{egbib}

\begin{thebibliography}{10}\itemsep=-1pt

\bibitem{Deeplabcrf}
L.-C. Chen, G.~Papandreou, I.~Kokkinos, K.~Murphy, and A.~L. Yuille.
\newblock Semantic image segmentation with deep convolutional nets and fully
  connected {CRFs}.
\newblock In {\em ICLR}, 2015.

\bibitem{Boxsup}
J.~Dai, K.~He, and J.~Sun.
\newblock Boxsup: Exploiting bounding boxes to supervise convolutional networks
  for semantic segmentation.
\newblock {\em arXiv preprint arXiv:1503.01640}, 2015.

\bibitem{DAICVPR15}
J.~Dai, K.~He, and J.~Sun.
\newblock Convolutional feature masking for joint object and stuff
  segmentation.
\newblock In {\em CVPR}, 2015.

\bibitem{Imagenet}
J.~Deng, W.~Dong, R.~Socher, L.-J. Li, K.~Li, and L.~Fei-Fei.
\newblock Imagenet: A large-scale hierarchical image database.
\newblock In {\em CVPR}, 2009.

\bibitem{Pascalvoc}
M.~Everingham, L.~Van~Gool, C.~K. Williams, J.~Winn, and A.~Zisserman.
\newblock The pascal visual object classes (voc) challenge.
\newblock {\em IJCV}, 88(2):303--338, 2010.

\bibitem{Farabet}
C.~Farabet, C.~Couprie, L.~Najman, and Y.~LeCun.
\newblock Learning hierarchical features for scene labeling.
\newblock {\em TPAMI}, 35(8):1915--1929, 2013.

\bibitem{Rcnn}
R.~Girshick, J.~Donahue, T.~Darrell, and J.~Malik.
\newblock Rich feature hierarchies for accurate object detection and semantic
  segmentation.
\newblock In {\em CVPR}, 2014.

\bibitem{Hariharan}
B.~Hariharan, P.~Arbel{\'a}ez, L.~Bourdev, S.~Maji, and J.~Malik.
\newblock Semantic contours from inverse detectors.
\newblock In {\em ICCV}, 2011.

\bibitem{Sds}
B.~Hariharan, P.~Arbel{\'a}ez, R.~Girshick, and J.~Malik.
\newblock Simultaneous detection and segmentation.
\newblock In {\em ECCV}, 2014.

\bibitem{Hypercolumns}
B.~Hariharan, P.~Arbel{\'a}ez, R.~Girshick, and J.~Malik.
\newblock Hypercolumns for object segmentation and fine-grained localization.
\newblock In {\em CVPR}, 2015.

\bibitem{LOFFEARXIV15}
S.~Ioffe and C.~Szegedy.
\newblock Batch normalization: Accelerating deep network training by reducing
  internal covariate shift.
\newblock {\em arXiv preprint arXiv:1502.03167}, 2015.

\bibitem{JiTPAMI13}
S.~Ji, W.~Xu, M.~Yang, and K.~Yu.
\newblock {3D} convolutional neural networks for human action recognition.
\newblock {\em TPAMI}, 35(1):221--231, 2013.

\bibitem{caffearxiv}
Y.~Jia, E.~Shelhamer, J.~Donahue, S.~Karayev, J.~Long, R.~Girshick,
  S.~Guadarrama, and T.~Darrell.
\newblock Caffe: Convolutional architecture for fast feature embedding.
\newblock {\em arXiv preprint arXiv:1408.5093}, 2014.

\bibitem{Fullycrf}
P.~Kr{\"a}henb{\"u}hl and V.~Koltun.
\newblock Efficient inference in fully connected crfs with gaussian edge
  potentials.
\newblock In {\em NIPS}, 2011.

\bibitem{Alexnet}
A.~Krizhevsky, I.~Sutskever, and G.~E. Hinton.
\newblock {ImageNet} classification with deep convolutional neural networks.
\newblock In {\em NIPS}, 2012.

\bibitem{LiACCV14}
S.~Li and A.~B. Chan.
\newblock {3D} human pose estimation from monocular images with deep
  convolutional neural network.
\newblock In {\em ACCV}, 2014.

\bibitem{Fcn}
J.~Long, E.~Shelhamer, and T.~Darrell.
\newblock Fully convolutional networks for semantic segmentation.
\newblock In {\em CVPR}, 2015.

\bibitem{Zoomout}
M.~Mostajabi, P.~Yadollahpour, and G.~Shakhnarovich.
\newblock Feedforward semantic segmentation with zoom-out features.
\newblock {\em arXiv preprint arXiv:1412.0774}, 2014.

\bibitem{Weaklyandsemi}
G.~Papandreou, L.-C. Chen, K.~Murphy, and A.~L. Yuille.
\newblock Weakly-and semi-supervised learning of a {DCNN} for semantic image
  segmentation.
\newblock {\em arXiv preprint arXiv:1502.02734}, 2015.

\bibitem{PinheiroArxiv2015}
P.~O. Pinheiro and R.~Collobert.
\newblock Weakly supervised semantic segmentation with convolutional networks.
\newblock In {\em CVPR}, 2015.

\bibitem{SimonyanNIPS2014}
K.~Simonyan and A.~Zisserman.
\newblock Two-stream convolutional networks for action recognition in videos.
\newblock In {\em NIPS}, 2014.

\bibitem{Vgg16}
K.~Simonyan and A.~Zisserman.
\newblock Very deep convolutional networks for large-scale image recognition.
\newblock In {\em ICLR}, 2015.

\bibitem{Googlenet}
C.~Szegedy, W.~Liu, Y.~Jia, P.~Sermanet, S.~Reed, D.~Anguelov, D.~Erhan,
  V.~Vanhoucke, and A.~Rabinovich.
\newblock Going deeper with convolutions.
\newblock {\em arXiv preprint arXiv:1409.4842}, 2014.

\bibitem{Visandund}
M.~D. Zeiler and R.~Fergus.
\newblock Visualizing and understanding convolutional networks.
\newblock In {\em ECCV}, 2014.

\bibitem{Deconvnet}
M.~D. Zeiler, G.~W. Taylor, and R.~Fergus.
\newblock Adaptive deconvolutional networks for mid and high level feature
  learning.
\newblock In {\em ICCV}, 2011.

\bibitem{Edgebox}
C.~L. Zitnick and P.~Doll{\'a}r.
\newblock Edge boxes: Locating object proposals from edges.
\newblock In {\em ECCV}, 2014.

\end{thebibliography}
}

\end{document}